\documentclass[conf]{new-aiaa}
\let\showhyphens\relax
\providecommand\showhyphens[1]{}
\usepackage[utf8]{inputenc}

\usepackage{graphicx}
\usepackage{amsmath}

\usepackage{amssymb,amsfonts}
\usepackage{algorithmic}
\usepackage{textcomp}
\usepackage{caption}
\usepackage{bm}
\usepackage[table]{xcolor}
\usepackage{longtable,tabularx}
\usepackage{siunitx}
\DeclareSIUnit\feet{ft}
\DeclareSIUnit{\nauticalmile}{nmi}
\DeclareSIUnit{\knot}{kn}

\hypersetup{
  bookmarksopen=true,
  colorlinks=true,
  linkcolor=black,
  citecolor=black,
  urlcolor=black,
}

\usepackage[acronym, nonumberlist, nopostdot, nomain]{glossaries}
\glsdisablehyper

\def\eg{e.g.\@}

\def\etc{etc.\@}
\def\ie{i.e.\@}

\newacronym{aam}{AAM}{Advanced Air Mobility}
\newacronym{faa}{FAA}{Federal Aviation Administration}
\newacronym{hrrr}{HRRR}{High-Resolution Rapid Refresh}
\newacronym{nurbs}{NURBS}{non-uniform rational B-spline}
\newacronym{fms}{FMS}{flight management system}
\newacronym{lpa}{LPA}{Light Propagation Algorithm}
\newacronym{gpr}{GPR}{Gaussian Process Regression}
\newacronym{bfs}{BFS}{Breadth-First Search}
\newacronym{fpv}{FPV}{flight plan validation}
\newacronym{rta}{RTA}{required time of arrival}
\newacronym{eta}{ETA}{estimated time of arrival}
\newacronym{swim}{SWIM}{System Wide Information Management}
\newacronym{adsb}{ADS-B}{Automatic Dependent Surveillance-Broadcast}
\newacronym{ot}{OT}{operational threshold}

\title{Uncertainty-Aware Conflict Detection Against Operator-Conditioned Weather Hazards}

\author{
    Balram Kandoria\footnote{Autonomy Engineer, SkyGrid.},
    Seulki Kim\textsuperscript{*},
    and Aryaman Singh Samyal\textsuperscript{*}
    }
\affil{SkyGrid, Austin, TX 78727, USA}

\begin{document}

\maketitle

\begin{abstract}
Strategic flight plan validation in Advanced Air Mobility (AAM) environments requires robust methods for predicting aircraft state uncertainty and detecting potential conflicts with dynamic airspace hazards. This paper presents a novel framework for uncertainty-conditioned trajectory prediction combined with polyhedra hazard representation for pre-flight conflict detection. We introduce a closed-form uncertainty estimation method that couples non-uniform rational B-spline (NURBS) curve fitting for kinematic trajectory generation with a Kalman Filter for state covariance propagation. Drawing from the Light Propagation Algorithm (LPA) paradigm, we employ a sigmoid-blended measurement noise model that captures the uncertainty reduction behavior of flight management systems approaching the required time of arrival (RTA) for waypoints. The resulting temporal uncertainty bounds are derived through a velocity-to-time variance transformation, enabling probabilistic assessment of arrival time deviations along the flight path. For hazard representation, we develop an operator-conditioned classification scheme that transforms gridded environmental data, specifically weather phenomena, into three-dimensional polyhedra volumes with intensity-based stratification. These hazard polyhedra incorporate aircraft-specific safety buffers computed from vehicle performance characteristics. Conflict detection is performed through mesh intersection algorithms operating on the spatial uncertainty tube surrounding the mean trajectory against the hazard polyhedra and temporal overlap. The framework enables the continuous strategic validation of flight plans throughout the pre-flight planning time horizon as environmental conditions evolve.
\end{abstract}

\section{Introduction}
\label{sec:introduction}

\subsection{Motivation}
\label{ssec:motivation}
The emergence of \gls{aam} and the increasing integration of unmanned aerial systems into the national airspace system present novel challenges for ensuring safe and efficient flight operations~\cite{kopardekar2016unmanned,thipphavong2018urban,national2020advancing}. Unlike traditional air traffic management paradigms that rely heavily on procedural separation and oversight, autonomous and highly automated vehicles operating in complex, dynamic environments require a framework for validating flight plans against evolving hazards. The validation must occur strategically, prior to departure, while accounting for the inherent uncertainty in predicting an aircraft's future state.

\Gls{fpv} serves as a critical function in the \gls{aam} operational ecosystem, assessing whether a proposed flight intent can be executed safely given the dynamic aeronautical and environmental conditions expected during the mission. These conditions encompass diverse weather phenomena (\eg, convective activity, icing, turbulence). The challenge lies not merely in determining whether a nominal trajectory intersects a hazard volume, but in quantifying the probability of conflict given the uncertainties inherent in the trajectory prediction and hazard characterization.

Among these uncertainties, temporal uncertainty is particularly consequential for \gls{aam} operations. A typical \gls{aam} mission spans 15--45~minutes, and cumulative arrival time uncertainty of 2--3~minutes represents 5--15\% of total flight duration, which is an order of magnitude larger than the equivalent ratio for multi-hour commercial flights. This uncertainty operates on the same timescale as transient low-altitude weather phenomena such as fog formation and dissipation cycles, passing convective cells, and intermittent wind gusts. The \gls{aam} operations in constrained urban corridors offer limited spatial rerouting flexibility, making temporal overlap---whether the aircraft arrives while a hazard is still active---the primary determinant of conflict. Without an adaptive pilot to respond in real time, and under UTM/U-space regulatory frameworks that mandate pre-departure strategic deconfliction~\cite{kopardekar2016unmanned,sesar2017uspace,astm2021utm}, robust temporal uncertainty quantification transitions from a theoretical refinement to an operational necessity.

\subsection{Problem Statement}
\label{ssec:problem_statement}
The \gls{fpv} problem can be formulated as follows. Consider an aircraft operating under a flight plan consisting of waypoints $\mathbf{w}_k = (x_k, y_k, z_k)$ with associated required times of arrival $t_k^{\text{RTA}}$ for $k = 0, 1, \ldots, N$. The aircraft's actual position at any time $t$ is subject to uncertainty arising from multiple sources, as characterized by performance-based navigation standards~\cite{icao2013pbn}:

\begin{enumerate}
    \item \textbf{Sensor measurement noise} in position and velocity estimation
    \item \textbf{Process uncertainty} from unmodeled dynamics and disturbances
    \item \textbf{Atmospheric factors} including wind, turbulence, and air density variations
    \item \textbf{Control system limitations} in tracking the commanded trajectory
\end{enumerate}

On the hazard side, weather hazards are characterized by spatial extent, temporal validity windows, and intensity levels. Since tolerance to a given weather phenomenon varies by aircraft type and operator equipage, hazard severity must be conditioned on operator-specific thresholds rather than treated as universally defined.

The \gls{fpv} problem thus requires:
\begin{itemize}
    \item Propagating trajectory uncertainty forward in time to establish probabilistic bounds on aircraft state;
    \item Representing hazards as three-dimensional volumes conditioned on operator-specific tolerance thresholds;
    \item Determining whether the uncertainty-conditioned trajectory conflicts with hazard volumes within specified confidence bounds.
\end{itemize}

\subsection{Related Work}
\label{ssec:related_work}

\subsubsection{Uncertainty Propagation in Trajectory Prediction}
\label{sssec:uncertinaty_propagation}
The problem of uncertainty quantification in aircraft trajectory prediction has been approached through various methodologies. Paielli and Erzberger~\cite{paielli1997conflict} introduced conflict probability estimation for free flight, employing a linear uncertainty model with along-track growth rates of 0.25 nautical miles per minute. While providing useful intuition about uncertainty propagation, this approach yields unbounded growth unsuitable for longer-duration missions.

Dougui et al.~\cite{dougui2012aircraft} advanced the state of the art with the \gls{lpa}, which models the uncertainty reduction behavior of \gls{fms}. Their approach assumes that uncertainty grows linearly until two-thirds of the required time of arrival has elapsed, after which the \gls{fms} compensates by adjusting airspeed to achieve an \gls{rta} tolerance of approximately 10 seconds. This bounded uncertainty model better reflects operational reality but relies on fixed growth rates that do not adapt to varying flight conditions. De Smedt et al.~\cite{smedt2013controlled} analyzed the feasibility of controlled time of arrival operations, demonstrating that modern \gls{fms} implementations achieve \gls{rta} tolerances on the order of 10--30 seconds, which informs the measurement noise parameterization in our Kalman Filter model, which we introduce in Section~\ref{sec:uncertainty-conditioned_trajectory_generation}.

Banerjee and Corbetta~\cite{banerjee2021uncertainty} presented a method for uncertainty quantification of expected time-of-arrival in UAV flight trajectories, employing \gls{nurbs} curves for kinematic profile generation combined with error interval propagation. Their work established a principled approach for transforming velocity uncertainty into temporal uncertainty through the relationship:

\begin{equation}
\sigma_{t_k}^2 = \sum_{i=1}^{k} \left( \frac{\Delta s_i}{\|\mathbf{v}_i\|} \cdot \frac{\sigma_{v_i}}{\|\mathbf{v}_i\|} \right)^2
\label{eq:banerjee_variance}
\end{equation}

\noindent where $\Delta s_i$ is the distance between waypoints, $\mathbf{v}_i$ is the velocity vector, and $\sigma_{v_i}$ is the velocity uncertainty. However, their approach requires empirical flight data to establish velocity variances, limiting applicability to pre-flight planning scenarios.

Corbetta et al.~\cite{corbetta2022uncertainty} extended uncertainty propagation to pre-flight prediction of separation violations for unmanned aerial vehicles, modeling wind effects as Gaussian random variables to capture environmental disturbances in the trajectory forecast. Marinescu et al.~\cite{marinescu2022wind} demonstrated the use of \gls{gpr} for spatiotemporal wind field estimation from aircraft surveillance data, providing both mean wind predictions and associated covariances suitable for integration with state estimation frameworks.

\subsubsection{Conflict Detection}
\label{sssec:conflict_detection}
Kuchar and Yang~\cite{kuchar2002review} provided a comprehensive survey of conflict detection and resolution methods, categorizing approaches by their treatment of trajectory uncertainty---from deterministic nominal paths to fully probabilistic state representations. Their taxonomy established the foundation for comparing detection algorithms and informed the design choices in this work. Prandini et al.~\cite{prandini2002probabilistic} formalized the probabilistic conflict detection problem, establishing randomized algorithms for computing conflict probability between stochastic trajectories.

\subsubsection{Weather Hazard Characterization}
\label{sssec:hazard_characterization}
Weather hazard characterization in commercial aviation has focused primarily on convective phenomena (\ie, thunderstorms and severe turbulence) where a pilot's operational decision upon encounter is largely binary: avoid or proceed. Krozel et al.~\cite{krozel2006turn} treated convective weather as weighted grid-based obstacles for en-route planning, and Michalek and Balakrishnan~\cite{michalek2009identification} advanced this by converting weather avoidance fields into two-dimensional convex polygonal obstacles. Tafferner et al.~\cite{tafferner2009improvement} addressed the altitude dimension by constructing extruded polygon volumes for thunderstorm cells, which are two-dimensional hazard boundaries extended across upper and lower altitude limits. These efforts have contributed to the safe operation of commercial aircraft around hazardous convective weather. However, low-altitude \gls{aam} operations with lighter and smaller aircraft introduce additional weather considerations beyond convection. Phenomena such as wind, visibility degradation, and temperature extremes are not binary hazards. Rather, their severity and the corresponding operational decisions should be determined by the performance envelope of the specific aircraft type.

To address the need for aircraft-type-specific severity classification in these emerging operations, Jones and Ellenbogen~\cite{jones2025risk} proposed and demonstrated the operational benefit of treating weather severity as a calibrated spectrum rather than a binary condition. Zhou et al.~\cite{zhouprobabilistic} introduced a user-defined weather risk threshold that allows the acceptable hazard level to vary by operator. However, these approaches remain at the level of flow management or trajectory risk assessment. Neither constructs geometric hazard volumes from weather data nor defines a multi-level severity taxonomy tied to specific aircraft performance envelopes.

This work addresses these gaps by constructing three-dimensional polyhedra volumes from weather grids with a multi-level severity (\ie, advisory, caution, warning) conditioned on aircraft-type-specific operational thresholds. These hazard volumes are then integrated into a strategic conflict detection framework that propagates trajectory temporal uncertainty to determine whether a planned flight conflicts with active hazard volumes within specified confidence bounds.

\subsection{Contributions}
\label{ssec:contribution}
This paper makes the following contributions to the \gls{fpv} problem:

\begin{enumerate}
    \item \textbf{Uncertainty-Conditioned Trajectory Generation}: We present a closed-form method combining \gls{nurbs}-based kinematic trajectory generation with Kalman Filter uncertainty propagation. A novel sigmoid-blended measurement noise model captures the \gls{fms} uncertainty reduction behavior without requiring discrete switching between open-loop and closed-loop modes.

    \item \textbf{Velocity-to-Time Uncertainty Transformation}: We formalize the conversion of spatial (\ie, position and velocity) uncertainty into temporal (\ie, \gls{rta}) uncertainty, enabling probabilistic assessment of arrival time deviations that can be compared against weather hazard temporal validity windows.

    \item \textbf{Operator-Conditioned polyhedra Hazards}: We develop a framework for constructing three-dimensional hazard polyhedra from gridded weather data with intensity stratification based on operator-defined thresholds.

    \item \textbf{Safety Buffer Computation}: We derive aircraft-specific safety buffer formulations that account for maneuverability constraints, enabling the hazard polyhedra to incorporate vehicle performance characteristics in their spatial extent.

    \item \textbf{Integrated Conflict Detection}: We present a conflict detection framework operating on the intersection of trajectory uncertainty tubes with hazard polyhedra to support strategic \gls{fpv}.
\end{enumerate}

\subsection{Paper Organization}
\label{ssec:paper_organization}
The remainder of this paper is organized as follows. Section~\ref{sec:uncertainty-conditioned_trajectory_generation} presents the mathematical framework for uncertainty-conditioned trajectory generation. Section~\ref{sec:polyhedra_hazards} develops the operator-conditioned hazard representation methodology. Section~\ref{sec:conflict_detection} describes the conflict detection algorithm. Section~\ref{sec:results} presents experimental results. Section~\ref{sec:conclusion} concludes with directions for future work.

\section{Uncertainty-Conditioned Trajectory Generation}
\label{sec:uncertainty-conditioned_trajectory_generation}

\subsection{Kinematic Profile from NURBS Curves}
\label{ssec:kinematic_profile}
Given a flight plan defined by waypoints $\mathbf{w}_k \in \mathbb{R}^3$ with associated timestamps $t_k$ for $k = 0, \ldots, N$, we construct a continuous trajectory representation using \gls{nurbs} interpolation~\cite{piegl1997b}. The waypoints are first transformed from geodesic coordinates (latitude, longitude, altitude) to a local East-North-Up (ENU) frame centered at the origin waypoint:

\begin{equation}
\mathbf{p}_{\text{ENU}} = \mathbf{R}_{\text{ENU}}(\mathbf{p}_{\text{geo}} - \mathbf{p}_{\text{ref}})
\label{eq:enu_transform}
\end{equation}
where $\mathbf{R}_{\text{ENU}}$ is the rotation matrix accounting for Earth curvature at the reference point $\mathbf{p}_{\text{ref}}$.

The \gls{nurbs} curve $\mathbf{C}(u)$ of degree $d$ is fitted to the transformed waypoints, yielding a parametric representation that provides $C^{d-1}$ continuity. The velocity profile along the trajectory is obtained through differentiation:

\begin{equation}
\mathbf{v}(t) = \frac{d\mathbf{C}}{dt} = \frac{d\mathbf{C}}{du} \cdot \frac{du}{dt}
\label{eq:velocity_profile}
\end{equation}

\subsection{Kalman Filter Formulation}
\label{ssec:kalman_filter}
We model the aircraft state as a six-dimensional vector comprising position and velocity components:
\begin{equation}
\mathbf{x} = [x, y, z, v_x, v_y, v_z]^T
\label{eq:state_vector}
\end{equation}

The control input $\mathbf{u}_k$ is derived from the \gls{nurbs} velocity profile. At each discrete time step $t_k$, the commanded velocity is extracted from the differentiated trajectory:
\begin{equation}
\mathbf{u}_k = \mathbf{v}(t_k) = \left. \frac{d\mathbf{C}}{dt} \right|_{t=t_k}
\label{eq:control_input}
\end{equation}

The discrete-time dynamics follow a constant-velocity model with acceleration noise:
\begin{equation}
\mathbf{x}_{k+1} = \mathbf{A}\mathbf{x}_k + \mathbf{B}\mathbf{u}_k + \mathbf{w}_k
\label{eq:dynamics}
\end{equation}
where the state transition matrix is:
\begin{equation}
\mathbf{A} = \begin{bmatrix} \mathbf{I}_3 & \Delta t \cdot \mathbf{I}_3 \\ \mathbf{0}_3 & \mathbf{I}_3 \end{bmatrix}
\label{eq:state_transition}
\end{equation}
the control input matrix is:
\begin{equation}
\mathbf{B} = \begin{bmatrix} \Delta t \cdot \mathbf{I}_3 \\ \mathbf{0}_3 \end{bmatrix}
\label{eq:control_matrix}
\end{equation}
and the process noise $\mathbf{w}_k \sim \mathcal{N}(\mathbf{0}, \mathbf{Q})$ captures acceleration disturbances with covariance:
\begin{equation}
\mathbf{Q} = \sigma_a^2 \begin{bmatrix} \frac{\Delta t^4}{4}\mathbf{I}_3 & \frac{\Delta t^3}{2}\mathbf{I}_3 \\ \frac{\Delta t^3}{2}\mathbf{I}_3 & \Delta t^2\mathbf{I}_3 \end{bmatrix}
\label{eq:process_noise}
\end{equation}
The measurement model observes position only:
\begin{equation}
\mathbf{z}_k = \mathbf{C}\mathbf{x}_k + \mathbf{v}_k
\label{eq:measurement_model}
\end{equation}
where $\mathbf{C} = [\mathbf{I}_3 \; \mathbf{0}_3]$ and $\mathbf{v}_k \sim \mathcal{N}(\mathbf{0}, \mathbf{R})$.

The filter is initialized with the first waypoint position and zero velocity, with initial covariance reflecting GPS and velocity estimation uncertainty:
\begin{equation}
\mathbf{x}_0 = [\mathbf{w}_0^T, \mathbf{0}_3^T]^T, \quad \mathbf{P}_0 = \begin{bmatrix} \sigma_{\text{pos}}^2 \mathbf{I}_3 & \mathbf{0}_3 \\ \mathbf{0}_3 & \sigma_{\text{vel}}^2 \mathbf{I}_3 \end{bmatrix}
\label{eq:initial_conditions}
\end{equation}
where typical values are $\sigma_{\text{pos}} = 5$ m (standard GPS error) and $\sigma_{\text{vel}} = 1$ m/s.

The standard Kalman Filter recursion~\cite{bar2001estimation} consists of the prediction step:
\begin{align}
\hat{\mathbf{x}}_{k|k-1} &= \mathbf{A}\hat{\mathbf{x}}_{k-1|k-1} + \mathbf{B}\mathbf{u}_{k-1} \label{eq:kf_predict_state}\\
\mathbf{P}_{k|k-1} &= \mathbf{A}\mathbf{P}_{k-1|k-1}\mathbf{A}^T + \mathbf{Q} \label{eq:kf_predict_cov}
\end{align}
and the update step:
\begin{align}
\mathbf{K}_k &= \mathbf{P}_{k|k-1}\mathbf{C}^T(\mathbf{C}\mathbf{P}_{k|k-1}\mathbf{C}^T + \mathbf{R})^{-1} \label{eq:kalman_gain}\\
\hat{\mathbf{x}}_{k|k} &= \hat{\mathbf{x}}_{k|k-1} + \mathbf{K}_k(\mathbf{z}_k - \mathbf{C}\hat{\mathbf{x}}_{k|k-1}) \label{eq:kf_update_state}\\
\mathbf{P}_{k|k} &= (\mathbf{I} - \mathbf{K}_k\mathbf{C})\mathbf{P}_{k|k-1} \label{eq:kf_update_cov}
\end{align}

\subsection{Sigmoid-Blended Measurement Noise Model}
\label{ssec:noise_model}
To capture the uncertainty reduction behavior of flight management systems without discrete switching, we introduce a progress-dependent measurement noise covariance:
\begin{equation}
\mathbf{R}(p) = \mathbf{R}_{\text{min}} + \frac{\mathbf{R}_{\text{max}} - \mathbf{R}_{\text{min}}}{1 + \exp(k(p - p_{\text{LPA}}))}
\label{eq:sigmoid_blend}
\end{equation}
where $p \in [0, 1]$ is the progress ratio between consecutive waypoints, $p_{\text{LPA}}$ is the activation threshold (typically $2/3$, following \gls{lpa} convention), and $k$ controls the blending sharpness. Prior art designated the $2/3$'s point between waypoints as the point where uncertainty convergence began or where \gls{fms} action would commence. In context of the Kalman Filter, this was implemented by not running the update step until after the activation point. Through experimentation we found the point which provides the smoothest blending is $p_{\text{LPA}}=0$ or when the update step is continuously running.

This formulation provides:
\begin{itemize}
    \item High measurement noise ($\mathbf{R} \approx \mathbf{R}_{\text{max}}$) early in each segment, yielding minimal Kalman gain and allowing uncertainty to grow
    \item Low measurement noise ($\mathbf{R} \approx \mathbf{R}_{\text{min}}$) as the aircraft approaches the waypoint, providing strong correction and uncertainty reduction
\end{itemize}

\begin{figure}[hbt!]
    \centering
    \includegraphics[width=0.8\columnwidth]{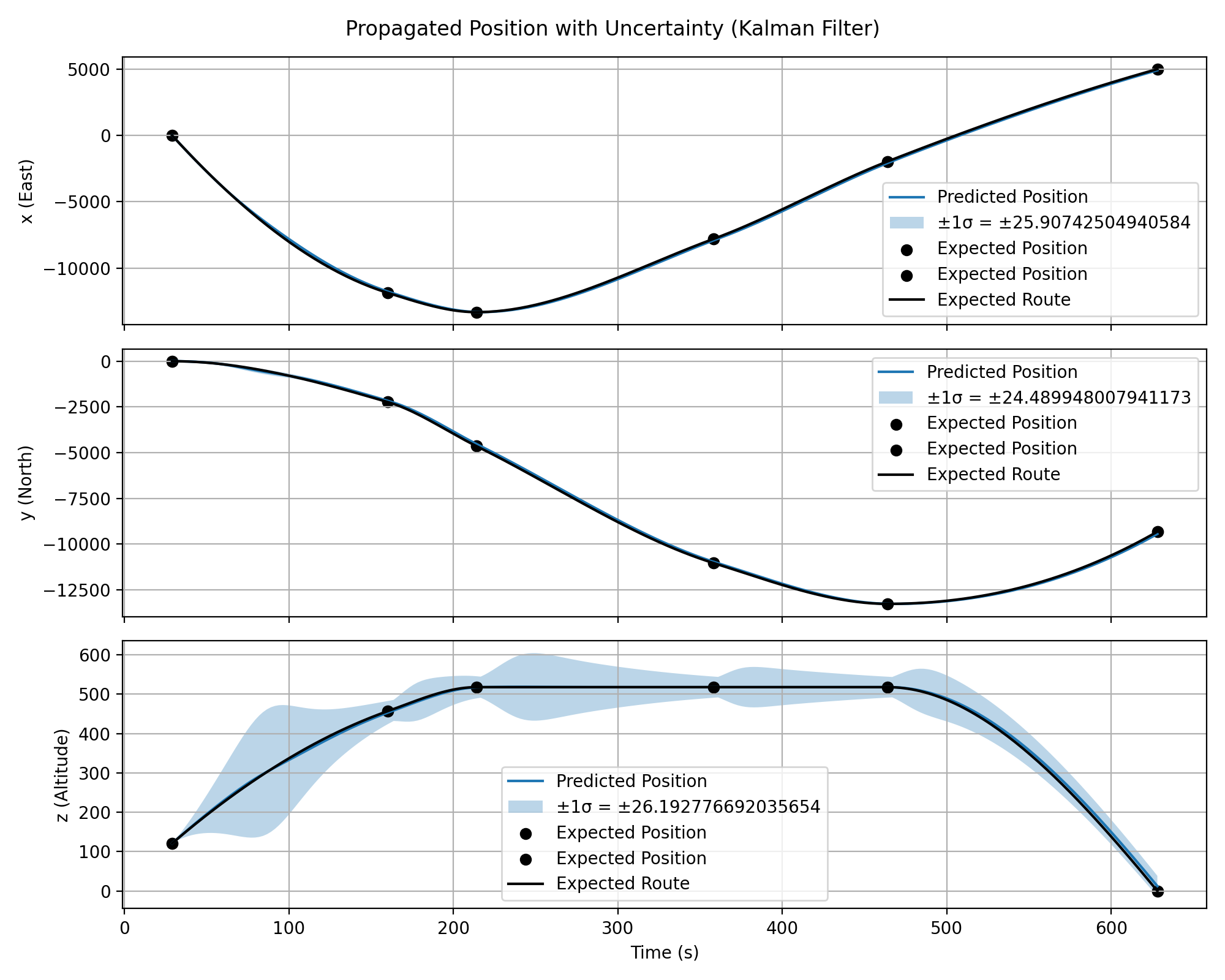}
    \caption{Kalman Filter with FMS simulated control behavior.}
    \label{fig:KF-FMS-SIMULATED-BEHAVIOR}
\end{figure}

The resulting uncertainty evolution naturally exhibits the bounded behavior characteristic of \gls{fms}-controlled flight without requiring explicit mode switching, see Fig.~\ref{fig:KF-FMS-SIMULATED-BEHAVIOR}. The setting point, $p_{\text{LPA}}=0$ indicates that the \gls{fms} is always active but progressively provides stronger control actions to meet \gls{rta} requirements.

\subsection{Temporal Uncertainty Derivation}
\label{ssec:temporal_uncertainty}
Following Banerjee and Corbetta~\cite{banerjee2021uncertainty}, we transform the propagated velocity uncertainty into temporal (\gls{rta}) uncertainty. At each point along the trajectory, the differential time based on velocity is:
\begin{equation}
\Delta t_i = \frac{\|\mathbf{p}_{i+1} - \mathbf{p}_i\|}{\|\mathbf{v}_i\|}
\label{eq:delta_t}
\end{equation}

The time variance contribution from each segment, subject to a minimum significance threshold, is:
\begin{equation}
\sigma_{t,i}^2 = \begin{cases} 0 & \text{if } \frac{\sigma_{v,i}}{\|\mathbf{v}_i\|} < \epsilon \\ \left(\Delta t_i \cdot \frac{\sigma_{v,i}}{\|\mathbf{v}_i\|}\right)^2 & \text{otherwise} \end{cases}
\label{eq:time_variance}
\end{equation}
where $\sigma_{v,i} = \|\bm{\sigma}_{\mathbf{v},i}\|$ is the velocity uncertainty magnitude extracted from the Kalman Filter covariance.

The cumulative \gls{rta} variance at waypoint $k$ is:
\begin{equation}
\sigma_{\text{RTA},k}^2 = \gamma^2 \sum_{i=1}^{k} \sigma_{t,i}^2
\label{eq:rta_variance}
\end{equation}
where $\gamma \leq 1$ is a safety factor that can be adjusted based on operational requirements.

\begin{figure}[hbt!]
    \centering
    \includegraphics[width=0.8\columnwidth]{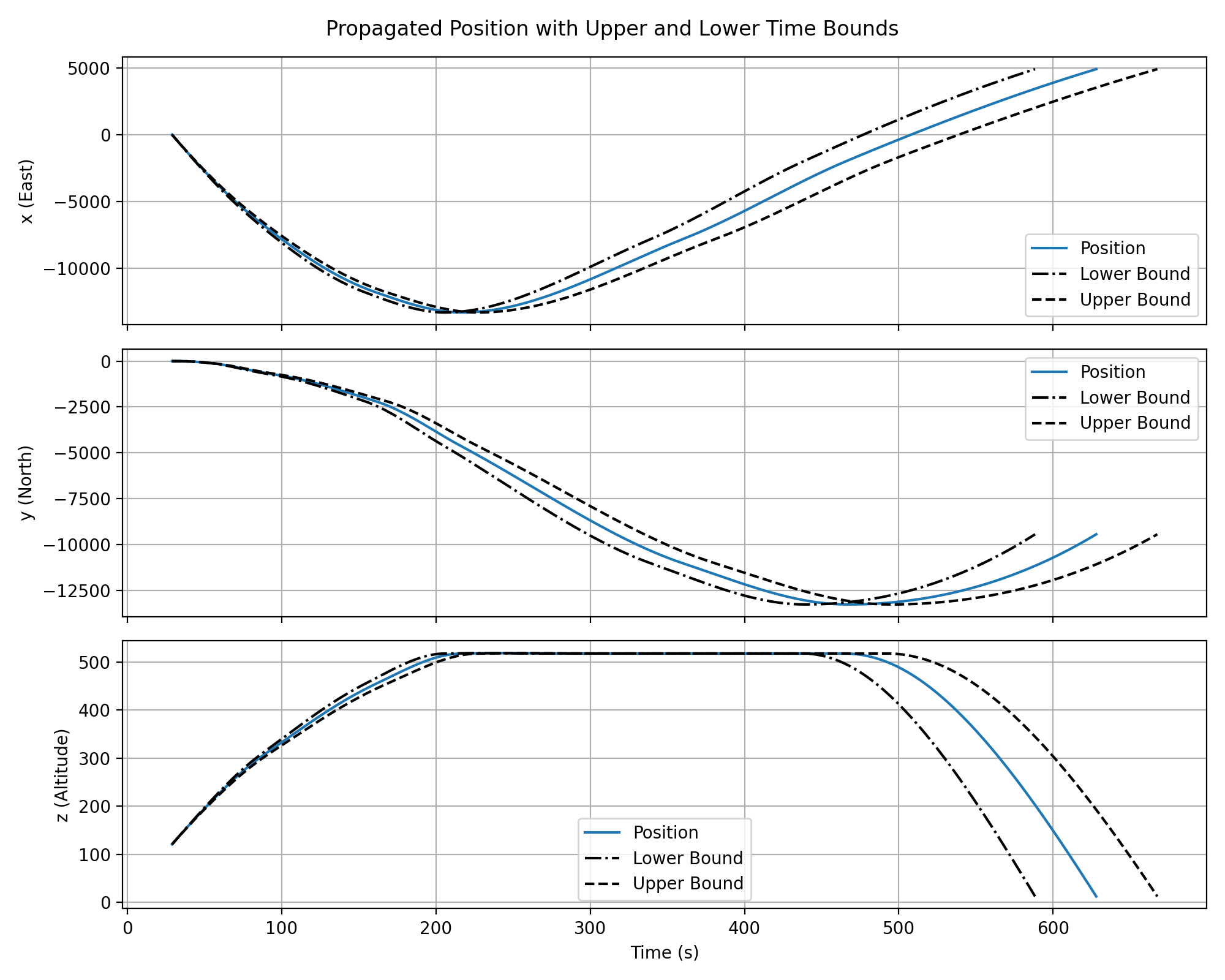}
    \caption{Upper and lower uncertainty bounds around a nominal trajectory.}
    \label{fig:upper_lower_uncertainty}
\end{figure}

The \gls{rta} at each waypoint can then be modeled as a Gaussian random variable:
\begin{equation}
t_k^{\text{RTA}} \sim \mathcal{N}(\bar{t}_k, \sigma_{\text{RTA},k}^2)
\label{eq:rta_distribution}
\end{equation}
enabling probabilistic bounds on arrival time for comparison against hazard temporal validity windows. A visual representation of the upper and lower bounds of the arrival time for a particular flight is shown in Fig.~\ref{fig:upper_lower_uncertainty}.
The variance is constantly increasing, as it is composed of the total sum of variance at a particular point.

\subsection{Wind Uncertainty Integration}
\label{ssec:wind_uncertinaty_integration}
External disturbances, particularly wind, significantly affect trajectory uncertainty. We incorporate wind effects through \gls{gpr}~\cite{williams2006gaussian,marinescu2022wind}, which provides both mean wind predictions and associated covariances along the flight route.

Given sparse wind observations $\{(\mathbf{p}_j, \mathbf{w}_j)\}_{j=1}^{M}$ where $\mathbf{w}_j = [u_j, v_j]^T$ represents eastward and northward wind components, \gls{gpr} fits a probabilistic model using a radial basis function kernel:
\begin{equation}
k(\mathbf{p}, \mathbf{p}') = \sigma_f^2 \exp\left(-\frac{\|\mathbf{p} - \mathbf{p}'\|^2}{2l^2}\right)
\label{eq:gpr_kernel}
\end{equation}
where $\sigma_f^2$ is the signal variance and $l$ is the length scale.

For any query point $\mathbf{p}^*$ along the trajectory, \gls{gpr} yields a predictive distribution:
\begin{equation}
\mathbf{w}(\mathbf{p}^*) \sim \mathcal{N}(\bm{\mu}_w(\mathbf{p}^*), \bm{\Sigma}_w(\mathbf{p}^*))
\label{eq:gpr_prediction}
\end{equation}

The wind vector is incorporated as an additive control input during propagation, and its covariance augments the process noise:
\begin{equation}
\mathbf{Q}_{wind} = \mathbf{Q} + \begin{bmatrix} \bm{\Sigma}_w & \bm{\Sigma}_w \\ \bm{\Sigma}_w & \bm{\Sigma}_w \end{bmatrix}
\label{eq:wind_augmented_noise}
\end{equation}
This formulation enables the uncertainty propagation to adapt to spatially varying wind conditions rather than assuming uniform disturbance characteristics.

\section{Operator-Conditioned polyhedra Hazards}
\label{sec:polyhedra_hazards}

\subsection{Hazard Classification}
\label{ssec:hazard_classification}
Weather hazards are classified against operator-defined \glspl{ot} that reflect
aircraft-specific tolerances and mission requirements. We consider temperature,
visibility, precipitation rate, and wind speed as weather attributes~$a$ in this
paper. For each attribute, three threshold levels are defined:
\begin{itemize}
    \item \textbf{\gls{ot}1 (Advisory)}: Conditions require increased awareness
    \item \textbf{\gls{ot}2 (Caution)}: Conditions approach performance limits
    \item \textbf{\gls{ot}3 (Warning)}: Conditions exceed safe operational limits
\end{itemize}

For a grid cell with indices $(i, j, k)$, the weather value $w_{i,j,k}^{(a)}$
for attribute~$a$ is compared against operator-defined thresholds. Let $\theta_\ell^{(a)}$
denote the operational threshold at level~$\ell \in \{1,2,3\}$ for attribute~$a$, where the levels correspond to Advisory, Caution, and Warning, respectively. Table~\ref{tab:operational_thresholds} presents representative operational
thresholds derived from \gls{faa} flight category definitions~\cite{faa2016weather}, WMO standard meteorological classifications~\cite{wmo2018guide}, and international aviation guidance~\cite{icao2018annex3}. These values serve as defaults when operator-specific thresholds are not provided. Table~\ref{tab:evtol_thresholds} presents the \gls{aam}-specific thresholds used in our experimental evaluation; however, it is worth noting that these thresholds are preliminary, representative values calibrated for a specific \gls{aam} aircraft profile. They will necessarily vary across vehicles depending on airframe design, propulsion architecture, \etc  ~Operators should establish platform-specific thresholds through flight testing and manufacturer guidance prior to operational deployment.
\begin{table}[hbt!]
\centering
\caption{Representative Operational Thresholds.}
\label{tab:operational_thresholds}
\begin{tabular}{lcccc}
\hline
\textbf{Attribute} & \textbf{Unit} & \textbf{OT1} ($\theta_1$) & \textbf{OT2} ($\theta_2$) & \textbf{OT3} ($\theta_3$) \\
\hline
Precipitation rate & mm/hr & 2.5 & 7.6 & 50.0 \\
Wind speed & kts & 20 & 35 & 50 \\
Visibility & SM & 5.0 & 3.0 & 1.0 \\
Temperature (max) & \si{\degreeCelsius} & 35 & 40 & 45 \\
Temperature (min) & \si{\degreeCelsius} & -10 & -20 & -30 \\
\hline
Separation buffer & NM & 15 & 20 & 30 \\
\hline
\end{tabular}
\end{table}
\begin{table}[hbt!]
\centering
\caption{AAM Operational Thresholds.}
\label{tab:evtol_thresholds}
\begin{tabular}{lcccc}
\hline
\textbf{Attribute} & \textbf{Unit} & \textbf{OT1} ($\theta_1$) & \textbf{OT2} ($\theta_2$) & \textbf{OT3} ($\theta_3$) \\
\hline
Precipitation rate & mm/hr & 1 & 2.5 & 7.6 \\
Wind speed & kts & 15 & 25 & 35 \\
Visibility & SM & 3.0 & 1.0 & 0.5 \\
Temperature (max) & \si{\degreeCelsius} & 30 & 35 & 45 \\
Temperature (min) & \si{\degreeCelsius} & 10 & 0 & -20 \\
\hline
Separation buffer & NM & 15 & 20 & 30 \\
\hline
\end{tabular}
\end{table}

Once weather values are compared against \glspl{ot}, each grid cell is assigned
a hazard level $H_{i,j,k}^{(a)}$ for attribute~$a$ as:
\begin{equation}
H_{i,j,k}^{(a)} = \begin{cases}
3 & \text{if } w_{i,j,k}^{(a)} \in \mathcal{D}_3^{(a)} \\
2 & \text{if } w_{i,j,k}^{(a)} \in \mathcal{D}_2^{(a)} \setminus \mathcal{D}_3^{(a)} \\
1 & \text{if } w_{i,j,k}^{(a)} \in \mathcal{D}_1^{(a)} \setminus \mathcal{D}_2^{(a)} \\
0 & \text{otherwise}
\end{cases}
\label{eq:hazard_level}
\end{equation}
where $\mathcal{D}_\ell^{(a)}$ denotes the hazardous domain at level $\ell$ for attribute $a$. For attributes where higher values indicate greater hazard severity
(\eg, precipitation rate, wind speed), the hazardous domain at level~$\ell$ is:
\begin{equation}
\mathcal{D}_\ell^{(a)} = \{w : w \geq \theta_\ell^{(a)}\}
\label{eq:hazard_domain_high}
\end{equation}
For attributes where lower values indicate greater hazard severity
(\eg, visibility):
\begin{equation}
\mathcal{D}_\ell^{(a)} = \{w : w \leq \theta_\ell^{(a)}\}
\label{eq:hazard_domain_low}
\end{equation}
By construction, $\theta_1^{(a)} \leq \theta_2^{(a)} \leq \theta_3^{(a)}$ for
higher-is-hazardous attributes (and reversed for lower-is-hazardous), ensuring
the nesting property $\mathcal{D}_3^{(a)} \subseteq \mathcal{D}_2^{(a)}
\subseteq \mathcal{D}_1^{(a)}$.

In practice, a grid cell may simultaneously exceed thresholds across multiple
weather attributes---for example, a cell may exhibit both high wind speed and
low visibility. To capture such compound conditions, we define the aggregate
hazard level as:
\begin{equation}
H_{i,j,k}^{(\mathrm{agg})} = \max_a H_{i,j,k}^{(a)}
\label{eq:aggregate_hazard}
\end{equation}
The max operator ensures that each cell is governed by its most severe
individual attribute. From an operational perspective, it is critical for operators to identify compounded-risk regions where multiple attributes simultaneously breach their respective thresholds. For instance, low visibility paired with strong winds
poses a significantly greater threat than either condition in isolation, and such regions must be avoided during \gls{fpv}.

\subsection{Spatial Clustering}
\label{ssec:spatial_clustering}

\subsubsection{Single-Attribute Clustering}
\label{sssec:single_clustering}
For each weather attribute, hazard cells are clustered together when they are
spatially adjacent and share the same hazard intensity level. We use a
\gls{bfs} algorithm with 26-neighborhood connectivity (including edge- and
corner-sharing neighbors) in the 3D grid. For a grid cell $c = (i,j,k) \in \Omega$, where $\Omega$ denotes the region-of-interest grid, let $\mathcal{N}_{26}(c)$ denote its 26-neighborhood of the cell $c$. For each attribute~$a$ and hazard level~$\ell$, the $m$-th spatially connected cluster is defined as
\begin{equation}
\mathcal{C}_{a,\ell}^{(m)} = \bigl\{c \in \Omega
  \;\big|\; H_{c}^{(a)} = \ell \text{ and connected under }
  \mathcal{N}_{26}\bigr\},
\label{eq:single_cluster}
\end{equation}
and serves as the basis for polyhedra construction.

\subsubsection{Aggregate Clustering}
\label{sssec:aggregate_clustering}
For aggregate hazards involving multiple contributing attributes, stricter
grouping conditions are imposed. Let
$\mathcal{A}_c = \{a : H_c^{(a)} > 0\}$ denote the set of active attributes
at cell~$c$, and let
$\mathbf{h} = [h_a]_{a \in \mathcal{A}}$ denote the corresponding hazard
level vector where $h_a = H_c^{(a)}$ for each contributing attribute. Two
adjacent cells are grouped into the same aggregate cluster only if they share
both the same active attribute set and identical per-attribute hazard levels.
The $m$-th aggregate cluster is defined as
\begin{equation}
\mathcal{G}_{\mathcal{A},\mathbf{h}}^{(m)} = \bigl\{c \in \Omega
  \;\big|\; \mathcal{A}_{c} = \mathcal{A},\;
  H_{c}^{(a)} = h_a\;\forall a \in \mathcal{A},\;
  \text{connected under } \mathcal{N}_{26}\bigr\}
\label{eq:aggregate_cluster}
\end{equation}
For example, consider two adjacent cells~$c_1$ and~$c_2$ with
$\mathcal{A}_{c_1} = \mathcal{A}_{c_2} = \{\text{wind\_speed},\,
\text{visibility}\}$. If $\mathbf{h}_{c_1} = [2,\,1]$ and
$\mathbf{h}_{c_2} = [1,\,2]$, these cells are assigned to separate
clusters despite sharing the same active attributes and the same aggregate
level $H^{(\mathrm{agg})} = 2$, because their per-attribute hazard levels
differ. This ensures that regions with distinct hazard compositions are
treated separately in polyhedron generation, even when their aggregate
levels coincide.

\subsection{Polyhedra Construction}
\label{ssec:polyhdra_construction}
Each cluster, whether single-attribute $\mathcal{C}_{a,\ell}^{(m)}$ or
aggregate $\mathcal{G}_{\mathcal{A},\mathbf{h}}^{(m)}$, is converted to a convex polyhedra volume. Let $(x_c, y_c, z_c)$ denote the geographic coordinates (longitude, latitude, altitude) of cell~$c$. For computational efficiency, we employ a vertical extrusion technique~\cite{de2008computational}:
\begin{enumerate}
    \item Project cluster cells onto the horizontal plane: $\mathcal{C}_{2D} = \{(x_c, y_c) : c \in \mathcal{C}\}$
    \item Construct 2D convex hull:           $\mathcal{H}_{2D} = \operatorname{ConvexHull}(\mathcal{P}_{2D})$
    \item Extract altitude bounds from the 3D cluster:
    \begin{align}
    z_{\min} &= \min_{c \in \mathcal{C}} z_c \label{eq:z_min} \\
    z_{\max} &= \max_{c \in \mathcal{C}} z_c \label{eq:z_max}
    \end{align}
\end{enumerate}
The resulting polyhedron is the lateral boundary $\mathcal{H}_{2D}$ extruded
vertically over $[z_{\min},\, z_{\max}]$, defined by vertices $\mathbf{V} = \{\mathbf{v}_1, \ldots, \mathbf{v}_M\}$ and triangular
faces~$\mathbf{F}$.

\subsection{Temporal Validity and Hazard Metadata}
\label{ssec:temporal_validity_hazard_metadata}
Each hazard polyhedron carries temporal validity that specifies when the hazard is active. For weather-derived hazards, validity windows are determined by the forecast timestamp of the underlying weather model:
\begin{equation}
\mathcal{W}_{\text{temporal}} = [t^{\text{start}}, t^{\text{end}}]
\label{eq:temporal_validity}
\end{equation}

Weather models such as the \gls{hrrr}~\cite{benjamin2016north} provide forecasts at discrete time steps (typically hourly), and the validity window extends from one forecast timestamp to the next. For NOTAMs and other regulatory hazards, explicit start and end times are provided in the hazard definition.

The complete hazard representation comprises:
\begin{itemize}
    \item Geometric extent: polyhedron vertices $\mathbf{V}$ and faces $\mathbf{F}$
    \item Altitude bounds: $[z_{\text{min}}, z_{\text{max}}]$
    \item Temporal validity: $[t^{\text{start}}, t^{\text{end}}]$
    \item Intensity level: $H \in \{1, 2, 3\}$
    \item Hazard type: single-attribute or aggregate
\end{itemize}
This metadata enables the conflict detection algorithm to filter hazards based on both spatial and temporal overlap with the trajectory.

\subsection{Aircraft-Specific Safety Buffers}
\label{ssec:safety_buffers}
Hazard polyhedra are expanded by safety buffers that account for aircraft maneuverability~\cite{eurocontrol2019bada}. The lateral safety radius~$r^*$ is determined by aircraft category.

\textbf{Fixed-wing aircraft} (performing turn-based avoidance):
\begin{equation}
r^*_{\text{FW}} = \frac{V_{\text{GS}}^2}{g \tan(\phi_{\text{max}})}
\label{eq:safety_range_fw}
\end{equation}
where $V_{GS}$ is ground speed, $g$ is gravitational acceleration, and $\phi_{\text{max}}$ is maximum bank angle.

\textbf{Rotary-wing aircraft} (performing deceleration-based avoidance):
\begin{equation}
r^*_{\text{RW}} = \frac{V_{\text{GS}}^2}{2a_{\text{max}}}
\label{eq:safety_range_rw}
\end{equation}
where $a_{\text{max}}$ is maximum deceleration rate.

For each triangular face~$f$ of the polyhedron with outward unit
normal~$\hat{\mathbf{n}}_f = (n_x, n_y, n_z)$, every vertex of~$f$ is
displaced along~$\hat{\mathbf{n}}_f$ with anisotropic scaling: $r^*$ for the
lateral components and a symmetric vertical buffer~$\Delta z_b$ for the
vertical component. The expanded polyhedron is then used and discretized into a triangular mesh $\mathcal{P}_{\text{hazard}}$ for downstream conflict detection.

\section{Conflict Detection}
\label{sec:conflict_detection}

\subsection{Trajectory Uncertainty Tube}
\label{ssec:trajectory_uncertainty_tube}
The spatial uncertainty from the Kalman Filter is represented as a three-dimensional tube surrounding the mean trajectory. Let $\bm{\mu}(t)$ denote the mean position at time~$t$ and
$r_{\mathrm{LoS}}$ a fixed loss-of-separation radius derived from the
position covariance. The uncertainty tube~$\mathcal{T}$ is defined as:
\begin{equation}
\mathcal{T} = \bigcup_{t} \{\mathbf{p} : \|\mathbf{p} - \bm{\mu}(t)\| \leq r_{\text{LoS}}\}
\label{eq:uncertainty_tube}
\end{equation}
The tube is then discretized into a triangular mesh with vertices $\mathbf{V}_T$ and faces $\mathbf{F}_T$.

\subsection{Mesh Intersection Algorithm}
\label{ssec:mesh_intersection}
Conflict detection is performed by computing the Boolean intersection of the trajectory tube mesh with each hazard polyhedron mesh~\cite{zhou2016mesh}:
\begin{equation}
\mathcal{I} = \mathcal{T} \cap \mathcal{P}_{\text{hazard}}
\label{eq:mesh_intersection}
\end{equation}
If the intersection mesh $\mathcal{I}$ has non-zero volume, a conflict exists. The intersection points identify the spatial extent of the conflict region.

\subsection{Temporal Conflict Assessment}
\label{ssec:temporal_conflict_assessment}
For segments where spatial intersection is detected, temporal overlap is assessed by comparing the \gls{rta} bounds with hazard validity windows:
\begin{equation}
\text{Conflict} \iff
\bigl[t_k^{\text{RTA}} - n\sigma_{\text{RTA},k},\; t_{k+1}^{\text{RTA}} + n\sigma_{\text{RTA},k+1}\bigr]
\cap
\bigl[t_{\text{hazard}}^{\text{start}},\, t_{\text{hazard}}^{\text{end}}\bigr] \neq \emptyset
\label{eq:temporal_conflict}
\end{equation}

\noindent where $n$ is the number of standard deviations for the desired confidence level.

\section{Experiments}
\label{sec:results}

\subsection{Trajectory Forecast Uncertainty Validation}
\label{ssec:kf_validation}
This experiment evaluates whether the Kalman Filter with sigmoid-blended measurement noise in Eq.~\eqref{eq:sigmoid_blend} accurately captures the uncertainty in trajectory forecasting when process noise is present. Validation was performed on real-flight data to demonstrate the viability of the Kalman Filter-based \gls{fms} control actions. \gls{faa} flight plan data was pulled from the \gls{faa} \gls{swim} database and the associated \gls{adsb} data was pulled from FlightRadar24. Table~\ref{tab:flight_times} delineates the flight plans parsed and used to validate whether modeled control actions could capture real total flight time.
\begin{table}[hbt!]
\centering
\caption{Flight time comparison between \gls{adsb} surveillance data and FAA flight plans.}
\label{tab:flight_times}
\begin{tabular}{|l|l|c|c|r|r|}
\hline
Callsign & Type & Origin & Dest & ADS-B Time & FAA Time \\
\hline\hline
AAL1943 & B738 & RDU & LAX & 4:43:49 & 4:46:00 \\
\hline
AAL2011\textsuperscript{*} & A321 & IAH & CLT & 1:56:12 & 2:01:00 \\
\hline
AAL830 & A321 & CLT & LAS & 3:56:29 & 4:07:00 \\
\hline
AAL853 & A319 & MMSP & DFW & 1:44:58 & 1:54:00 \\
\hline
AAY1455 & A320 & VPS & MDW & 1:55:04 & 2:00:37 \\
\hline
AAY446 & A320 & DSM & BOS & 2:29:42 & 2:24:09 \\
\hline
AAY622 & A320 & IND & LAS & 3:27:42 & 3:33:17 \\
\hline
ARG1302 & A332 & SAEZ & MIA & 8:32:28 & 8:17:01 \\
\hline
DAL3041 & B712 & JAN & ATL & 0:57:11 & 1:05:00 \\
\hline
DAL923 & B752 & BOS & ATL & 2:08:47 & 2:13:00 \\
\hline
JBU1204 & A320 & MDSD & EWR & 3:35:47 & 3:34:17 \\
\hline
KLM757 & B77W & EHAM & MPTO & 10:11:48 & 10:05:07 \\
\hline
SWA1241 & B737 & PHX & SAT & 1:48:34 & 1:51:00 \\
\hline
SWA1494 & B737 & PDX & SMF & 1:05:56 & 1:13:00 \\
\hline
SWA2081 & B738 & MDW & PIT & 0:55:59 & 1:04:00 \\
\hline
SWA2269 & B738 & MDW & BUF & 1:02:59 & 1:12:00 \\
\hline
SWA2476 & B737 & GEG & PHX & 2:18:39 & 2:15:00 \\
\hline
SWA3461 & B738 & LAS & STL & 2:38:47 & 2:43:00 \\
\hline
SWA527 & B737 & ORD & DEN & 2:17:42 & 2:11:00 \\
\hline
SWA663 & B737 & AUS & MCO & 2:15:39 & 2:12:00 \\
\hline
SWA739 & B737 & STL & CMH & 1:07:36 & 1:03:00 \\
\hline
SWA9006 & B737 & DEN & LAX & 1:58:46 & 2:06:00 \\
\hline
UAL2617 & B738 & LGA & IAH & 2:56:00 & 3:02:00 \\
\hline
\end{tabular}
\vspace{0.5em}

\footnotesize{\textsuperscript{*}ADS-B track begins mid-flight at 10,550~ft; duration is measured from the first recorded point.}
\end{table}

The \gls{faa} flight plans were simulated using the Kalman Filter-based \gls{fms} uncertainty propagation method in order to extract the variance induced from trained unmodeled environment conditions. The resulting variance was applied to the total planned flight time, as specified in the \gls{faa} flight plan, in order to predict actual aircraft performance.

\begin{table}[hbt!]
\centering
\caption{Kalman Filter verification results across $n = 23$ flights (16 passed, 7 failed, accuracy = 69.6\%).}
\label{tab:kf_verification}
\begin{tabular}{|l|r|r|r|c|}
\hline
Callsign & ADS-B Time & FAA Time & $\hat{\sigma}_t$ (s) & Result \\
\hline\hline
AAL1943 & 4:43:49 & 4:46:00 & 858.2 & PASS \\
\hline
AAL2011 & 1:56:12 & 2:01:00 & 351.4 & PASS \\
\hline
AAL830 & 3:56:29 & 4:07:00 & 657.0 & PASS \\
\hline
AAL853 & 1:44:58 & 1:54:00 & 630.9 & PASS \\
\hline
AAY1455 & 1:55:04 & 2:00:37 & 360.4 & PASS \\
\hline
AAY446 & 2:29:42 & 2:24:09 & 420.1 & PASS \\
\hline
AAY622 & 3:27:42 & 3:33:17 & 537.6 & PASS \\
\hline
ARG1302 & 8:32:28 & 8:17:01 & 3042.7 & PASS \\
\hline
DAL3041 & 0:57:11 & 1:05:00 & 269.9 & FAIL \\
\hline
DAL923 & 2:08:47 & 2:13:00 & 357.9 & PASS \\
\hline
JBU1204 & 3:35:47 & 3:34:17 & 739.1 & PASS \\
\hline
KLM757 & 10:11:48 & 10:05:07 & 3435.7 & PASS \\
\hline
SWA1241 & 1:48:34 & 1:51:00 & 332.8 & PASS \\
\hline
SWA1494 & 1:05:56 & 1:13:00 & 274.1 & FAIL \\
\hline
SWA2081 & 0:55:59 & 1:04:00 & 231.6 & FAIL \\
\hline
SWA2269 & 1:02:59 & 1:12:00 & 259.0 & FAIL \\
\hline
SWA2476 & 2:18:39 & 2:15:00 & 372.5 & PASS \\
\hline
SWA3461 & 2:38:47 & 2:43:00 & 401.8 & PASS \\
\hline
SWA527 & 2:17:42 & 2:11:00 & 364.7 & FAIL \\
\hline
SWA663 & 2:15:39 & 2:12:00 & 344.5 & PASS \\
\hline
SWA739 & 1:07:36 & 1:03:00 & 222.7 & FAIL \\
\hline
SWA9006 & 1:58:46 & 2:06:00 & 395.1 & FAIL \\
\hline
UAL2617 & 2:56:00 & 3:02:00 & 482.6 & PASS \\
\hline
\end{tabular}
\end{table}

Approximately 70\% of simulations were able to predict real-aircraft performance with the Kalman Filter-\gls{fms} method without having specific knowledge of the flight environment. Simulations where the Kalman Filter-based \gls{fms} method did not predict real-flight dynamics accurately can be attributed to tactical deviations from the filed flight plan. It is expected that \gls{aam} flights will fly autonomously and thus have stricter adherence to their planned trajectory making the Kalman Filter-\gls{fms} method more reliable.

\subsection{Uncertainty-Conditioned Conflict Detection}
\label{ssec:uncertainty_conflict_detection}
This experiment demonstrates the practical impact of incorporating temporal uncertainty into \gls{fpv}. We construct scenarios where a flight route through the Houston airspace spatially intersects weather hazards, but the hazards have temporal validity windows that do \textit{not} overlap with the deterministic \gls{eta}. When the \gls{rta} uncertainty from the Kalman Filter is incorporated in Eq.~\eqref{eq:temporal_conflict}, the expanded time window reveals conflicts that deterministic validation misses.

For a 30-minute \gls{aam} mission with $\sigma_{\text{RTA}}$ of 2--3~minutes, the $2\sigma$ arrival window spans approximately 10~minutes. This is comparable to the active lifetime of low-altitude weather phenomena such as fog dissipation or convective cell passage, meaning that a deterministic \gls{eta} may nominally miss a hazard by a margin well within the \gls{rta} uncertainty. The following experiment quantifies this effect using realistic weather data.

\subsubsection{Experiment Design}
\label{sssec:experimental_design}
\begin{figure}[hbt!]
    \centering
    \includegraphics[width=0.5\columnwidth]{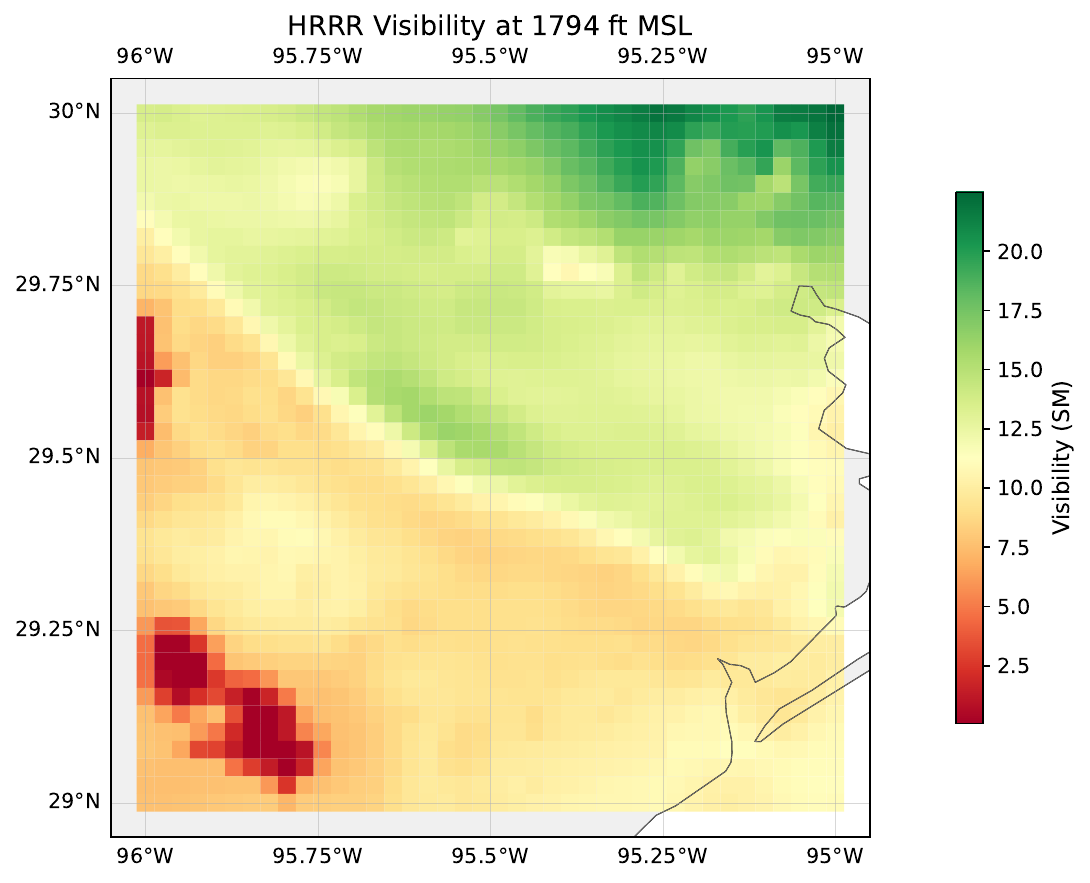}
    \caption{HRRR visibility over the Houston area at a low-altitude layer.}
    \label{fig:visibility_raw}
\end{figure}

To evaluate the conflict detection framework under realistic conditions, we generate weather hazards from \gls{hrrr} data~\cite{benjamin2016north} covering the Houston, Texas metropolitan area. The \gls{hrrr} provides gridded atmospheric data at 3~km horizontal resolution for attributes such as temperature, $u-$ and $v-$wind components, visibility, and precipitation rate. Figure~\ref{fig:visibility_raw} illustrates a visual example of visibility from the \gls{hrrr} sample at a representative low-altitude layer.

\begin{figure}[hbt!]
    \centering
    \includegraphics[width=0.8\columnwidth]{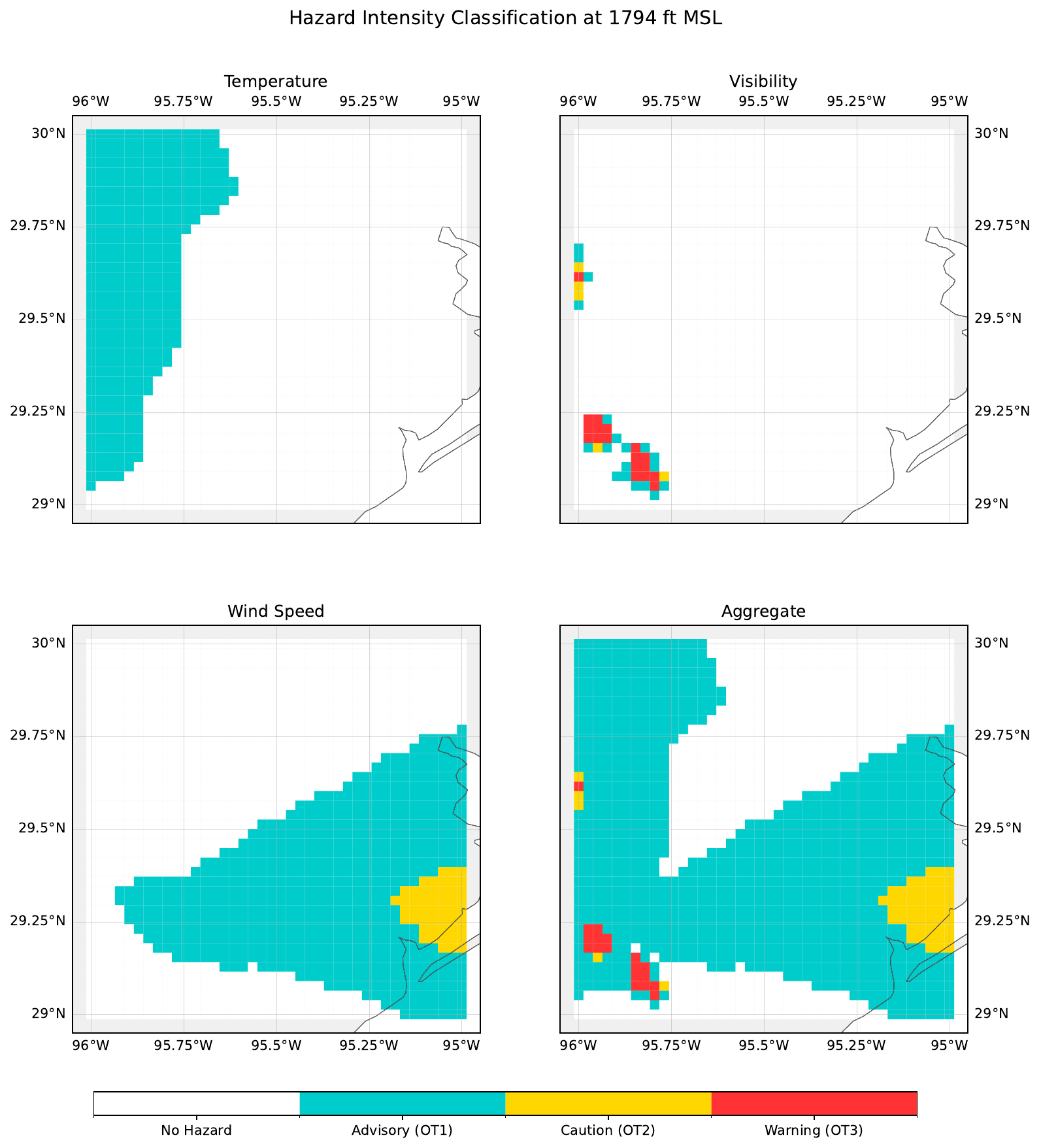}
    \caption{Hazard intensity classification at a low-altitude layer for single-attribute (temperature, visibility, and wind speed), and the aggregate hazard. Grid cells are classified as Advisory (OT1), Caution (OT2), or Warning (OT3) per the AAM thresholds in Table~\ref{tab:evtol_thresholds}.}
    \label{fig:hazard_intensity}
\end{figure}

The hazard generation pipeline processes the raw netCDF \gls{hrrr} data through four stages:
\begin{enumerate}
    \item \textbf{Data ingestion and unit conversion}: HRRR weather attributes are converted from native units to aviation-standard operational units (\eg, \si{K} $\to$  \si{\degreeCelsius}, m/s $\to$ kts, m $\to$ ft).
    \item \textbf{Hazard classification}: Each grid cell is evaluated against the \gls{aam} operational thresholds in Table~\ref{tab:evtol_thresholds} to produce single-attribute hazard intensity matrices $H_{i,j,k}^{(a)}$ and the aggregate hazard matrix $H_{i,j,k}^{(\mathrm{agg})}$.
    \item \textbf{Spatial clustering}: Hazardous cells are grouped via breadth-first search with 26-connectivity, yielding spatially coherent hazard volumes.
    \item \textbf{Polyhedra construction and buffer expansion}: Clusters are converted to convex polyhedra via vertical extrusion, then expanded by the aircraft-specific safety buffer $r^* \approx 787$~m, with $V_{\text{GS}} = 60$~m/s, $\phi_{\max} = 25^\circ$) laterally and $\pm 500$~ft vertically.
\end{enumerate}
Figure~\ref{fig:hazard_intensity} shows the hazard intensity classification for individual weather attributes and the aggregate at a representative altitude layer, illustrating how different weather phenomena contribute to the overall hazard landscape.

The Houston dataset yields 33 hazard polygons: 15 single-attribute and 18 aggregate hazards. Table~\ref{tab:houston_hazards} summarizes the distribution by intensity level and contributing phenomena.
\begin{table}[hbt!]
\centering
\caption{Weather Hazard Distribution over Houston.}
\label{tab:houston_hazards}
\begin{tabular}{lccc}
\hline
\textbf{Phenomenon} & \textbf{Advisory} & \textbf{Caution} & \textbf{Warning} \\
\hline
Temperature & 5 & -- & -- \\
Visibility & 5 & -- & 3 \\
Wind speed & 1 & 1 & -- \\
Aggregate (multi) & 13 & -- & 5 \\
\hline
\textbf{Total} & 24 & 1 & 8 \\
\hline
\end{tabular}
\end{table}

\begin{figure}[hbt!]
    \centering
    \includegraphics[width=1.0\columnwidth]{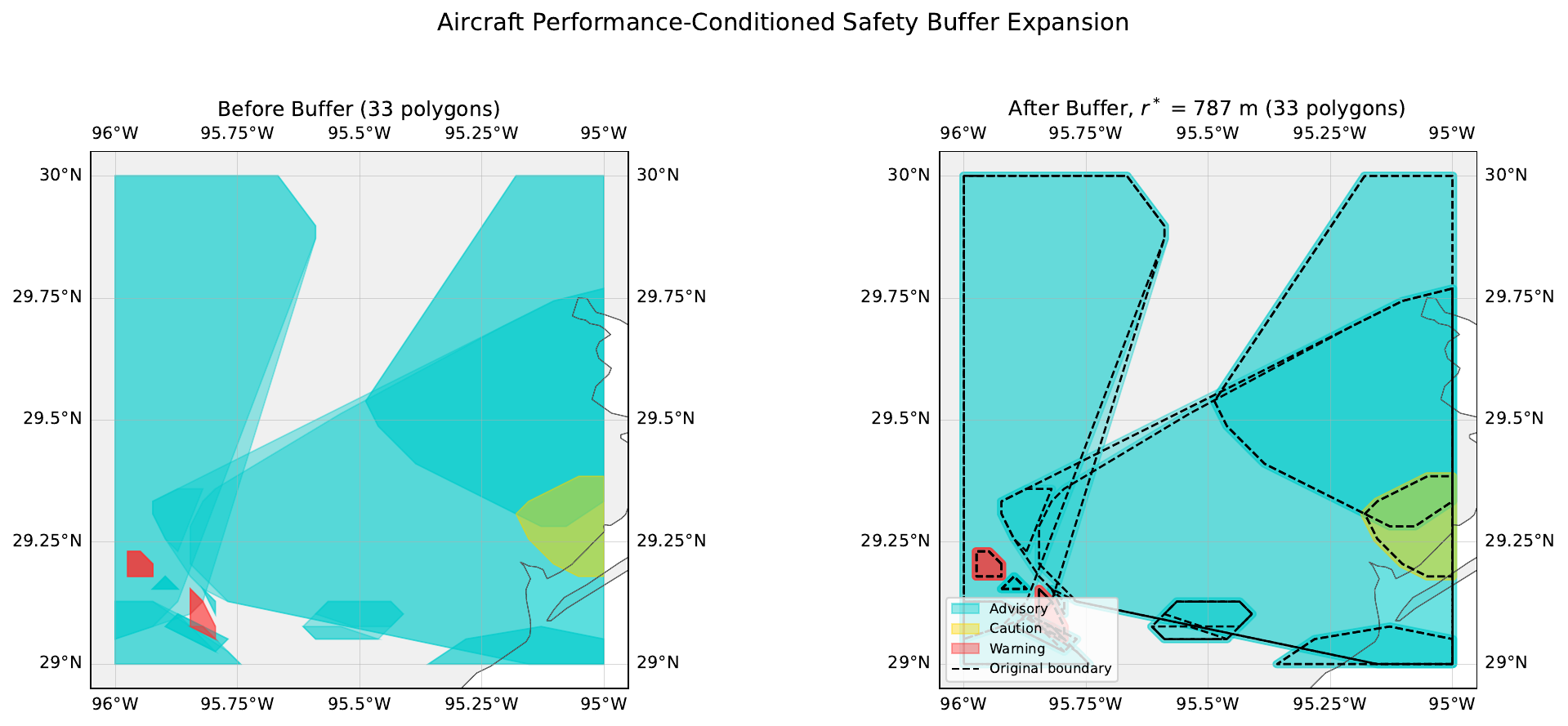}
    \caption{Houston weather hazard polygons colored by intensity (Advisory, Caution, Warning). Left: original projected polygonal boundaries. Right: after lateral safety buffer expansion ($r^* \approx 787$~m, Eq.~\eqref{eq:safety_range_fw}). Dashed black outlines show original boundaries for comparison.}
    \label{fig:buffer_comparison}
\end{figure}

Figure~\ref{fig:buffer_comparison} shows all 33 hazard polygons projected onto the Houston area, colored by intensity level, and illustrates the effect of the aircraft performance-conditioned safety buffer by comparing polygon extents before and after the $r^*$ lateral expansion. The dashed outlines on the right panel indicate the original boundaries prior to buffer application.
Each hazard is assigned a temporal validity window derived from the \gls{hrrr} forecast cycle. Visibility and wind hazards, which are inherently transient, have narrower windows (5--30~min active periods), while temperature-dominated hazards persist across the full forecast hour. Hazards with narrow validity windows are particularly susceptible to being missed by deterministic validation but caught when temporal uncertainty is incorporated.

A flight route is then defined through the Houston hazard field, traversing regions with active weather hazards at various altitudes. The trajectory is processed through the full \gls{fpv} pipeline:
\begin{enumerate}
    \item \textbf{Kinematic profile generation}: \gls{nurbs}-based path fitting with PCHIP time interpolation produces the nominal trajectory with 250 sample points.
    \item \textbf{Uncertainty propagation}: The Kalman Filter propagates state and covariance, with the sigmoid-blended measurement noise capturing \gls{fms} convergence behavior near waypoints.
    \item \textbf{Temporal uncertainty extraction}: Velocity uncertainty is converted to \gls{rta} variance via Eq.~\eqref{eq:time_variance}, yielding cumulative $\sigma_{\text{RTA}}$ at each waypoint.
    \item \textbf{Conflict detection}: Both deterministic (zero variance) and uncertainty-conditioned ($n\sigma$ bounds) validations are run against the same hazard set.
\end{enumerate}

\subsubsection{Results}
\label{sssec:results}
\begin{figure}[hbt!]
    \centering
    \includegraphics[width=0.7\columnwidth]{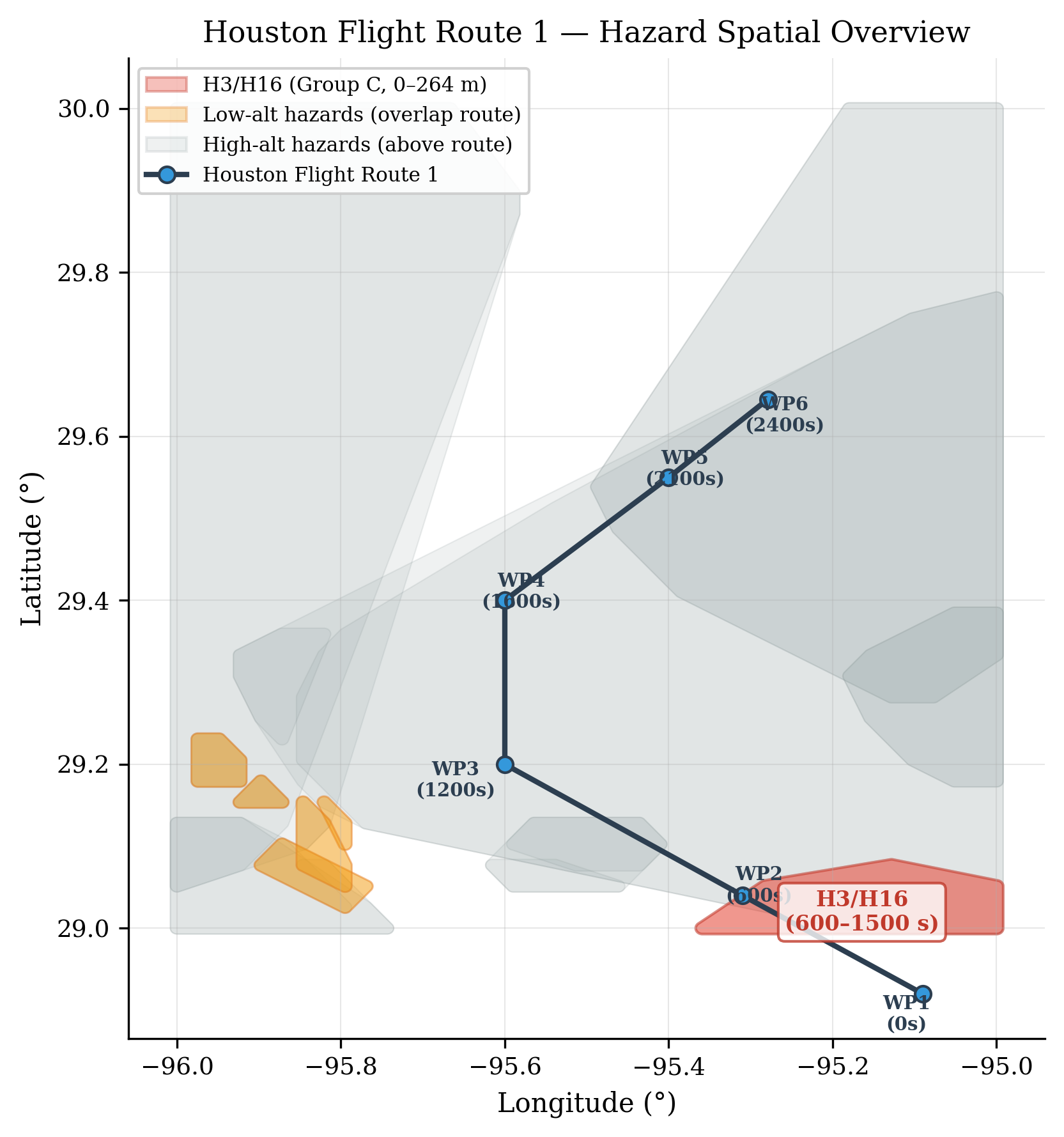}
    \caption{Houston Flight Route with Weather Hazards.}
    \label{fig:houston_route}
\end{figure}

A nominal trajectory to HOU was designed to cross hazards, but produce an FPV evaluation of ``no conflicts'', see the flight route shown in Fig.~\ref{fig:houston_route}.
The hazards in gray are excluded from the FPV simulation as their altitudes are above that expected from a typical \gls{aam} flight.

After running the strategic conflict detector and including the uncertainty variance derived from the Kalman Filter \gls{fms} approach, the \gls{fpv} check yielded a potential for a conflict during segment 0. A summary of the simulation results is stored in Table~\ref{tab:temporal_overlap}.
\begin{table}[hbt!]
\centering
\caption{Temporal overlap analysis for Houston Flight Route~1 against surviving hazard groups.}
\label{tab:temporal_overlap}
\small
\setlength{\tabcolsep}{5pt}
\begin{tabular}{|l|c|c|c|c|c|}
\hline
& \cellcolor{red!8} \textbf{Seg 0} & \textbf{Seg 1} & \textbf{Seg 2} & \textbf{Seg 3} & \textbf{Seg 4} \\
\hline
$t_{\mathrm{end}}$ (s) & \cellcolor{red!8} 600 & 1200 & 1600 & 2100 & 2400 \\
\hline
$T_{\mathrm{var}}$ (s) & \cellcolor{red!8} 139.34 & 209.11 & 248.57 & 301.11 & 328.88 \\
\hline
Nom.\ RTA (s) & \cellcolor{red!8} 600.0 & 1200.0 & 1600.0 & 2100.0 & 2400.0 \\
\hline
Unc.\ + RTA (s) & \cellcolor{red!8} 739.3 & 1409.1 & 1848.6 & 2401.1 & 2728.9 \\
\hline\hline
\textbf{A}: 0--1800\,s & \cellcolor{red!8} $\checkmark$/$\checkmark$ & $\checkmark$/$\checkmark$ & $\checkmark$/$\checkmark$ & $\checkmark$/$\checkmark$ & $\times$/$\times$ \\
\hline
\textbf{B}: 300--1080\,s & \cellcolor{red!8} $\checkmark$/$\checkmark$ & $\checkmark$/$\checkmark$ & $\times$/$\times$ & $\times$/$\times$ & $\times$/$\times$ \\
\hline
\textbf{C}: 600--1500\,s & \cellcolor{red!8} $\checkmark$/$\times$ & $\checkmark$/$\checkmark$ & $\checkmark$/$\checkmark$ & $\times$/$\times$ & $\times$/$\times$ \\
\hline
\textbf{D}: 1200--2400\,s & \cellcolor{red!8} $\times$/$\times$ & $\checkmark$/$\times$ & $\checkmark$/$\checkmark$ & $\checkmark$/$\checkmark$ & $\checkmark$/$\checkmark$ \\
\hline\hline
Result (Nom.) & \cellcolor{red!8} P$^\dagger$ & P$^\dagger$ & P$^\dagger$ & P$^\dagger$ & P$^\dagger$ \\
\hline
Result (Unc.) & \cellcolor{red!8} \textbf{F} & P$^\dagger$ & P$^\dagger$ & P$^\dagger$ & P$^\dagger$ \\
\hline
\end{tabular}
\end{table}

 Of the 33 hazards, 12 pass the first-pass altitude filter (route max alt.\ 60\,m + 100\,m buffer $< 173.3$\,m). These form four groups by active time window.
``$\checkmark$'' = temporal overlap, ``$\times$'' = no temporal overlap;
boldface \textbf{F} = confirmed 3D conflict, $\dagger$ = temporal overlap only (no 3D intersection).

Table~\ref{tab:temporal_overlap} shows all segments with passing nominal and uncertainty-infused \glspl{rta} except for segment 0. At approximately 600~s \gls{rta}, the aircraft sees an additional 139 seconds of time variance, shifting it into the active time of the associated hazard. Thus in the non-nominal case a conflict is predicted to occur purely based on the dynamics of the vehicle and the unmodeled characteristics of the flight environment.

\begin{figure}[hbt!]
    \centering
    \includegraphics[width=0.8\columnwidth]{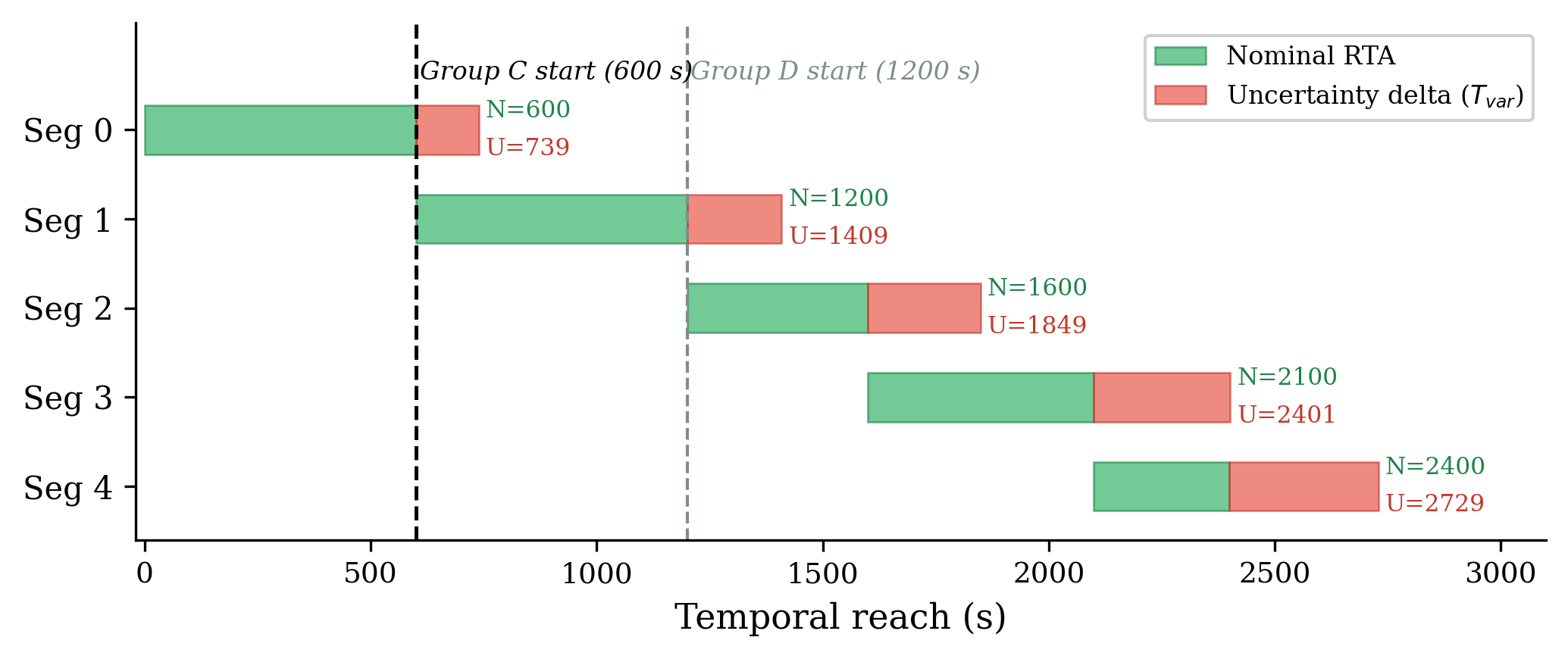}
    \caption{Time Overlap Analysis.}
    \label{fig:KF-FMS-TIME-ANALYSIS}
\end{figure}

Specifically, hazard group~C (active window 600--1500~s) is the discriminating case. The temporal analysis, Fig.~\ref{fig:KF-FMS-TIME-ANALYSIS}, depicts the start time boundary of the validity window~C, for the related hazard group, and the nominal \gls{rta} in segment~0, which is at t=600~s. Under deterministic validation, this yields no temporal overlap. However, when the cumulative \gls{rta} variance of 139~s is applied, the uncertainty-conditioned arrival time extends to 739~s, shown as the red bar extension on segment 0. This shift into the hazard's active window, with applied temporal deltas, triggers a confirmed 3D mesh intersection and thus a detected conflict. The remaining segments exhibit temporal overlap with at least one hazard group under both nominal and uncertainty-conditioned evaluation but produce no 3D spatial intersection since the maximum AAM flight ceiling is below that of the hazards lowest extrusion point, yielding passing results throughout.

This result illustrates the mechanism by which temporal uncertainty changes the conflict verdict: the deterministic \gls{eta} falls just outside a hazard's validity window by a margin smaller than the \gls{rta} standard deviation $\sigma_{\text{RTA}}$. In such cases, deterministic validation reports no conflict while uncertainty-conditioned validation correctly identifies a potential hazard encounter. A flight plan approved under deterministic assumptions may traverse active hazards when arrival time deviations---which grow cumulatively along the route---are properly accounted for. For \gls{aam} operations where 139~s of uncertainty on a 10-minute segment represents a 23.17\% deviation, this demonstrates that temporal uncertainty quantification is not a theoretical refinement but a practical necessity for safe pre-departure \gls{fpv}.

\section{Conclusion}
\label{sec:conclusion}
This paper presented a framework for uncertainty-aware strategic flight plan validation combining Kalman Filter-based trajectory uncertainty propagation, operator-conditioned polyhedra hazard generation from gridded weather data, and 3D mesh intersection conflict detection. The trajectory uncertainty model, validated against 23 real flights using \gls{faa} \gls{swim} / \gls{adsb} data, achieved a capture rate of 69.6\% without knowing the en-route condition -- a figure expected to improve in \gls{aam} operations where autonomous vehicles adhere more strictly to planned trajectories. The hazard generation pipeline produced 33 operator-conditioned polyhedra from \gls{hrrr} data over Houston, classified at three intensity levels with aircraft performance-conditioned safety buffers ($r^* \approx 787$~m) that scale avoidance margins to the operational vehicle.

The experimental result demonstrated the practical consequence of temporal uncertainty: a flight plan passing deterministic validation was found to conflict with a hazard when a cumulative \gls{rta} variance of 139~s--- 23.17\% of a 10-minute segment---shifted the arrival time into an active hazard window. This confirms that for \gls{aam} missions, where temporal uncertainty operates on the same timescale as transient low-altitude weather phenomena, deterministic validation alone is insufficient.

Current limitations include the constant-velocity process model (which omits climb, descent, and turning dynamics), convex hull overestimation for non-convex hazard clusters, and the single-route evaluation scope. Future work will incorporate higher-fidelity process models, non-convex hazard representations via alpha shapes, real-time weather observation feeds for dynamic hazard updating, and multi-flight scaling for UTM/U-space fleet-level deconfliction.

\bibliography{refs/references}

\end{document}